\documentclass[letterpaper, 10pt, conference]{ieeeconf}      

\IEEEoverridecommandlockouts 
\usepackage{cite}

\usepackage{amsmath,amssymb,amsfonts}
\usepackage{graphicx}
\usepackage{textcomp}

\DeclareMathOperator*{\argmin}{arg\,min}
\usepackage[table,xcdraw]{xcolor}
\usepackage{authblk}
\let\labelindent\relax
\usepackage{enumitem}
\usepackage{hyperref}
\usepackage{breqn}
\usepackage{algpseudocode}
\usepackage[linesnumbered,ruled,lined]{algorithm2e}
\usepackage{subfigure}

\usepackage{array}
\usepackage{makecell}
\usepackage{tabularray}

\usepackage{svg}

\allowdisplaybreaks

\begin{document}

\title{ End-to-End Learning of Behavioural Inputs for Autonomous Driving in Dense Traffic }
\author{Jatan Shrestha, Simon Idoko, Basant Sharma, Arun Kumar Singh\thanks{All authors are with the University of Tartu. This research was in part supported by financed by European Social Fund via ICT program measure, grants PSG753 from Estonian Research Council and collaboration project LLTAT21278 with Bolt Technologies. Our code is available at \url{https://github.com/jatan12/DiffProj.git}.
Emails: jatanshrestha7@gmail.com, aks1812@gmail.com} 
}
\maketitle


\begin{abstract} 
Trajectory sampling in the Frenet(road-aligned) frame, is one of the most popular methods for motion planning of autonomous vehicles. It operates by sampling a set of behavioral inputs, such as lane offset and forward speed, before solving a trajectory optimization problem conditioned on the sampled inputs. The sampling is handcrafted based on simple heuristics, does not adapt to driving scenarios, and is oblivious to the capabilities of downstream trajectory planners. 


In this paper, we propose an end-to-end learning of behavioral input distribution from expert demonstrations or in a self-supervised manner. We embed a novel differentiable trajectory optimizer as a layer in neural networks, allowing us to update behavioral inputs by considering the optimizer's feedback. Moreover, our end-to-end approach also ensures that the learned behavioral inputs aid the convergence of the optimizer. We improve the state-of-the-art in the following aspects. First, we show that learned behavioral inputs substantially decrease collision rate while improving driving efficiency over handcrafted approaches. Second, our approach outperforms model predictive control methods based on sampling-based optimization.


\end{abstract}

\section{Introduction} 
The planning layer for autonomous driving includes two hierarchical components. At the top level, the behavioral layer computes decisions such as lane change, speeding up, and braking based on the traffic scenario and the driving task. The behavioral inputs can be parameterized as set points for longitudinal velocity, lateral offsets from the center line, and goal positions. Such representation naturally integrates with the downstream optimal trajectory planner \cite{li2021safe} \cite{wei2014behavioral}, \cite{lim2019hybrid}.


\subsubsection*{Existing Gaps} 
\begin{figure}[t!]
\centering
 \includegraphics[scale=0.330]{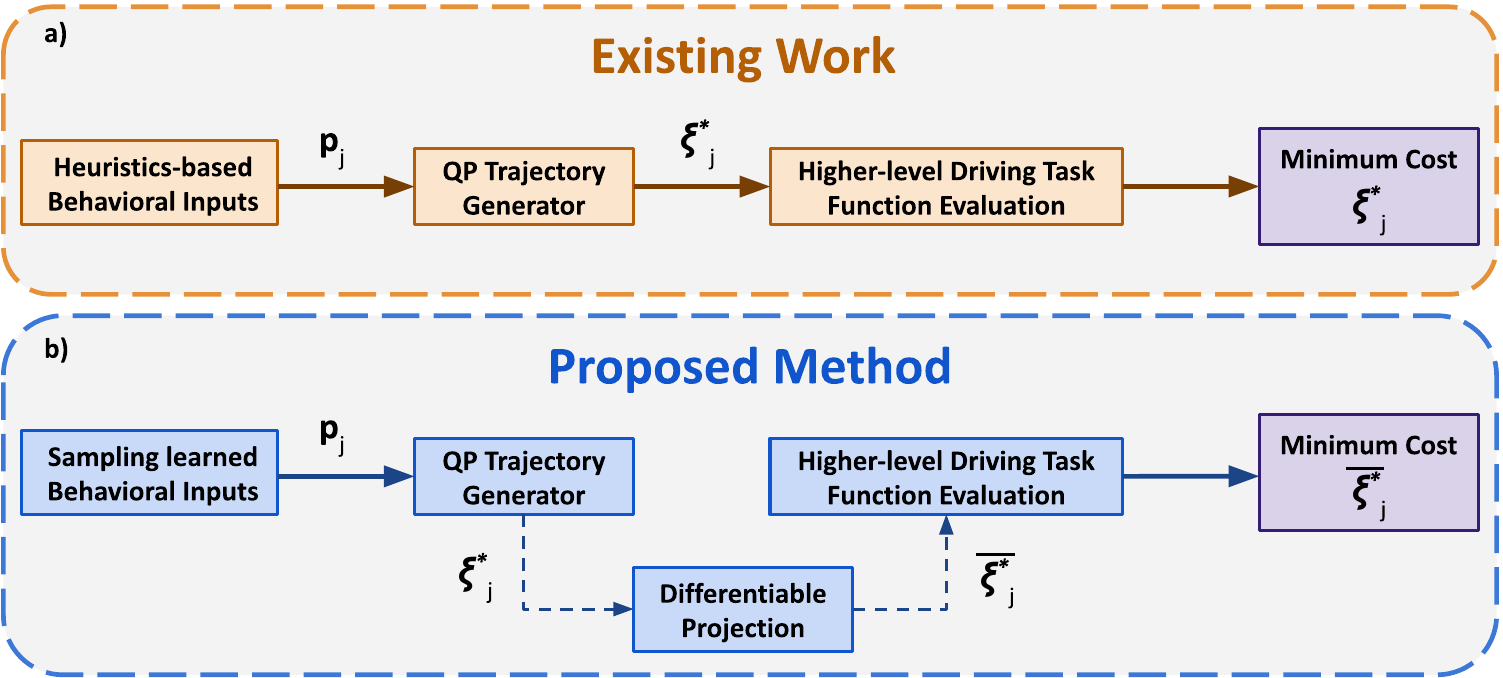}
\caption{Comparison between existing and proposed pipeline for motion planning in autonomous driving. We differ in two respects. First, we sample behavioral inputs from a learned distribution. Second, the downstream trajectory planner has a projection optimizer to aid in the satisfaction of collision and kinematic constraints. We also learn the optimizer initialization along with the behavioral inputs.}
\label{pipeline}
\vspace{-0.70cm}
\end{figure}

Existing approaches \cite{li2021safe}, \cite{wei2014behavioral}, \cite{werling2010optimal} for computing optimal behavior inputs and motion plans consist of two steps (see Fig.\ref{pipeline}). The behavioral inputs are sampled based on simple heuristics and then fed to the downstream trajectory optimizer. The resulting trajectories are then ranked based on their performance on the driving tasks, modeled through some meta costs- cruising speed, collision avoidance, etc.
 
There are three fundamental problems associated with the existing approaches. First, the behavioral input sampling is handcrafted, usually sampled from a pre-specified grid. Second, the sampling does not adapt to driving scenarios and the capabilities of downstream trajectory optimizers. Third, the planner itself is just a simple Quadratic Programme (QP) without explicit collision-avoidance and kinematic constraints.  

This paper presents an end-to-end learning method addressing the core problems discussed above. The end-to-end aspect of our approach signifies that we jointly learn behavioral inputs and initialization for our trajectory optimizer while considering their interactions (see Remark \ref{rem_1}). Our core innovations and their associated benefits are summarized below.

\noindent \subsubsection*{ Algorithmic Contribution} 
\begin{itemize}
    \item We propose a supervised and also self-supervised approach for learning behavioral inputs. For the former, we use a Conditional Variational Autoencoder(CVAE) \cite{NIPS2015_8d55a249} that directly learns a distribution over optimal behavioral inputs. For the latter, we use Multi-Layer Perceptron (MLP) and treat its output as the mean of the distribution.

    \item We propose a differentiable constrained optimizer that improves QP-based planning and can also be embedded as a layer in neural networks. The resulting backpropagation traces the gradient of the loss function through the differentiable constrained optimizer. We show that our optimizer has an efficient batchable structure and allows for the pre-storing of expensive computations such as matrix factorizations.
\end{itemize}

\noindent \subsubsection*{ State-of-the-Art Performance} 

\begin{itemize}
    \item Our end-to-end learning approach outperforms planning with handcrafted behavioural inputs (e.g. \cite{adajania2022multi}) in collision rate and achieved speed metrics. The performance gap increases as the traffic becomes dense.

    \item We also achieve a lower collision rate than Model Predictive Path Integral (MPPI), a state-of-the-art sampling-based optimizer.
\end{itemize}

\section{Mathematical Preliminaries}
\subsubsection*{Symbols and Notation} Normal font lower-case letters will represent scalars, and bold font variants will represent vectors. The upper-case bold font letters will represent matrices. The superscript $T$ will denote the transpose of a matrix or a vector.



\subsection{Frenet Frame and Trajectory Parametrization}

\noindent We formulate motion planning of the ego-vehicle in the road-aligned reference known as the Frenet frame. In this setting, the longitudinal and lateral motions of the ego-vehicle are always aligned with the $X$ and $Y$ axes of the Frenet-frame respectively. We parametrize the positional space ($x(t), y(t)$) of the ego-vehicle  in the Frenet frame at any time instant $t$ in terms of polynomials:

\vspace{-0.3cm}

\small
\begin{align}
    \begin{bmatrix}
        x(t_0), \dots, x(t_f) 
    \end{bmatrix} = \textbf{W}\textbf{c}_{x},
     \begin{bmatrix}
        y(t_0), \dots, y(t_f) 
    \end{bmatrix} = \textbf{W}\textbf{c}_{y},
    \label{param}
\end{align}
\normalsize

\noindent where, $\textbf{W}$ is a matrix formed with time-dependent polynomial basis functions and ($\textbf{c}_{x}, \textbf{c}_{y}$) are the coefficients of the polynomial. We can also express the derivatives in terms of $\dot{\textbf{W}}, \ddot{\textbf{W}}$.


\subsection{Behavioral Input Parametrization}
\noindent We summarize the commonly used behavioural inputs below

\begin{itemize}
    \item $\textbf{p}_d = (y_d, v_d)$: The desired lateral offset from the centre line and longitudinal speed.
    \item $\textbf{p}_{term} = (x_f, y_f, \dot{x}_f, \dot{y}_f, \ddot{x}_f, \ddot{y}_f)$: Final states along the longitudinal and lateral directions.
\end{itemize}
\noindent We stack all the behavioural inputs into one parameter vector:
\vspace{-0.3cm}
\small
\begin{align}
\textbf{p} = \begin{bmatrix}
   \textbf{p}_d, \textbf{p}_{term} 
\end{bmatrix}.
\label{behaviour_param}
\end{align}
\normalsize

\noindent Note that not all elements of $\textbf{p}$ need to be used simultaneously in the downstream planner. For example, \cite{hoel_rl_behavior}, use a single set-point for lateral offset and desired velocity as behavioural inputs while authors in \cite{adajania2022multi} uses only $(x_f, y_f)$. It is also possible to expand the list. For longer horizons, we can split the planning horizon segments into $m$ parts and assign individual lateral offsets $y_{d, m}$ and desired speed $v_{d, m}$ to each of these segments.


\subsection{Existing Behavioural and Trajectory Planning}

\subsubsection{Downstream Trajectory Planner}
\noindent We can obtain different formulations for the trajectory planner, depending on the choice of behavioural inputs. We present below a generic construction that draws inspiration from \cite{li2021safe}, \cite{wei2014behavioral}, \cite{werling2010optimal}, \cite{adajania2022multi} and work for all the behavioural inputs presented in the previous subsection.

\vspace{-0.5cm}
\small
\begin{subequations}
\begin{align}
  \min  \sum_t c_{s} +c_{l}+c_v\label{cost} \\
    (x^{(r)}(t_0),  y^{(r)}(t_0)) = \textbf{b}_0, (x^{(r)}(t_f),  y^{(r)}(t_f) ) = \textbf{p}_{term} \label{boundary_cond}
\end{align}
\end{subequations}
\normalsize
\vspace{-0.5cm}
\small
\begin{subequations}
\begin{align}
    c_{s} (\ddot{x}(t), \ddot{y}(t)) = \ddot{x}(t)^2+\ddot{y}(t)^2\\
    c_{l}(\ddot{y}(t), \dot{y}(t)) = (\ddot{y}(t)-\kappa_p(y(t)-y_d)-\kappa_v\dot{y}(t))^2\\
    c_v(\dot{x}(t), \ddot{x}(t)) = (\ddot{x}(t)-\kappa_p(\dot{x}(t)-v_d))^2
\end{align}
\end{subequations}
\normalsize

\noindent The first term $c_s(.)$ in the cost function \eqref{cost} ensures smoothness in the planned trajectory by penalizing high accelerations at discrete time instants. The last two terms ($c_l(.), c_v(.)$) model the tracking of lateral offset ($y_{d}$) and forward velocity $(v_{d})$ set-points respectively and is inspired from works like \cite{hoel_rl_behavior}. For the former, we define a Proportional Derivative (PD) like tracking with gain $(\kappa_p, \kappa_v)$. It induces lateral accelerations that will make the ego-vehicle converge to the $y_d$. The derivative terms in $c_l$ minimize oscillations while converging to the desired lateral offset. For velocity tracking, we only use a proportional term.  Equality constraints \eqref{boundary_cond} ensures boundary conditions on  the $r^{th}$ derivative of the planned trajectory. We use $r= \{0, 1, 2\}$ in our formulation.

Optimization \eqref{cost}-\eqref{boundary_cond} is a convex QP. To make this form more explicit, we can use the parametrization proposed in \eqref{param} to put the above optimization into a more compact form
\small
\begin{subequations}
\begin{align}
    \argmin_{\boldsymbol{\xi}} \frac{1}{2}\boldsymbol{\xi}^T\textbf{Q}\boldsymbol{\xi}+\textbf{q}^T(\textbf{p})\boldsymbol{\xi}, \label{lower_cost_reform}  \\
    \textbf{A}\boldsymbol{\xi} = \textbf{b}(\textbf{p})\label{lower_eq_reform} 
\end{align}
\end{subequations}
\normalsize
\noindent where $\boldsymbol{\xi} = (\textbf{c}_x, \textbf{c}_y)$. A part of $\textbf{p}$ that models lateral offsets and desired velocity enters the cost function
while the rest enters the r.h.s of the equality constraints.

\subsubsection{Sampling and meta Cost} Let $\textbf{p}_j$ be the $j^{th}$ behavioral input sampled from a fixed distribution (or a grid). Existing works like \cite{li2021safe}, \cite{wei2014behavioral}, \cite{werling2010optimal}, \cite{adajania2022multi} solve \eqref{lower_cost_reform}-\eqref{lower_eq_reform} for all $\textbf{p}_j$ and rank the resulting trajectories based on some higher-level (meta) cost function. Let  $\boldsymbol{\xi}_j^* = (\textbf{c}_x^*, \textbf{c}_y^*)$ be the resulting optimal trajectory coefficients corresponding to $\textbf{p}_j$. Accordingly, the meta cost used in this work to model the driving task can be defined as follows:

\vspace{-0.4cm}

\small
\begin{align}
    c_{m} (\boldsymbol{\xi}) =  c_{res}(\boldsymbol{\xi}^*)+\left \Vert \dot{\textbf{W}}\textbf{c}_x^*-v_{des}\right\Vert_2^2,
    \label{meta_cost}
\end{align}
\normalsize
\noindent where $c_{res}$ measures the residual (violation) of kinematic and collision avoidance constraints. The second term in $c_m$ measures the deviation from some desired longitudinal speed. In dense traffic scenarios, a heuristic sampling of $\textbf{p}_j$ is likely to lead to a high meta-cost for all trajectories. In the next section, we introduce our main result; replacing the hand-crafted sampling with a neural network trained in an end-to-end fashion.

\begin{figure*}[htp]
    \centering
    \includegraphics[scale=0.30]{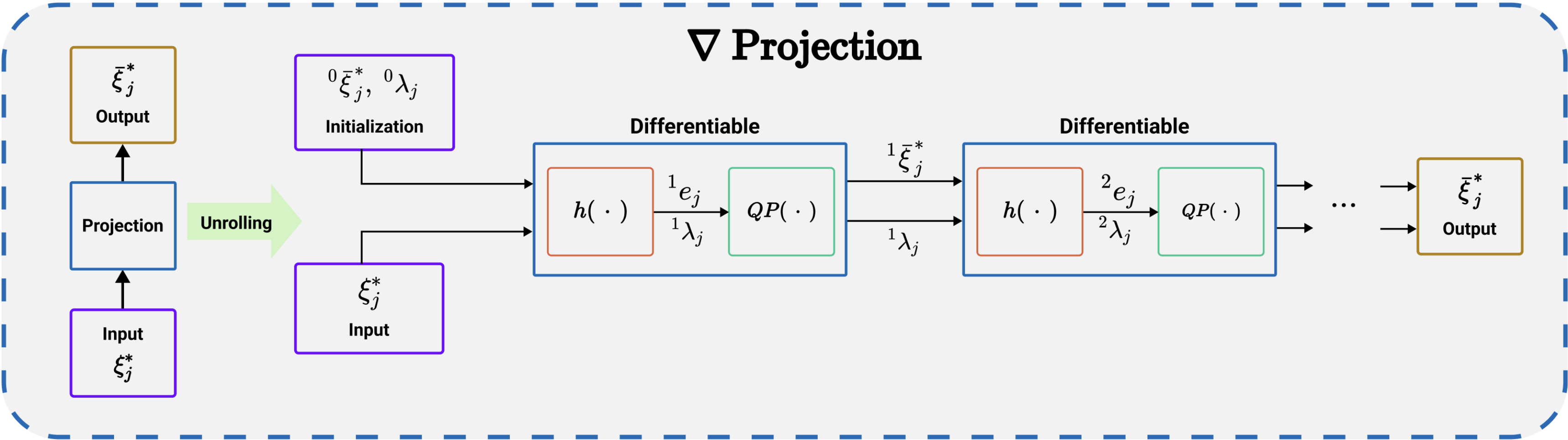}
    \caption{Unrolled representation of our projection optimizer formed by a repeated stacking of an analytical function $\textbf{h}(.)$ and a $QP(.)$ block. We can compute the gradient of any loss function defined on the output $\overline{\boldsymbol{\xi}}_j^*$ with respect to any intermediate value of the unrolling pipeline. }
    \label{unrolling}
    \vspace{-0.5cm}
\end{figure*}

\section{Main Results}
Fig.\ref{pipeline}(b) presents an overview of our main algorithmic results that have two key components. First, our trajectory planner consists of QP \eqref{lower_cost_reform}-\eqref{lower_eq_reform} augmented with a differentiable projection module. Second, the behavioural inputs are sampled from a learned distribution. We present the first component next.

\subsection{Differentiable Constrained Optimizer}
\noindent Our projection optimizer has the following form
\small
\begin{align}
    \overline{\boldsymbol{{\xi}}}^{*}_j = \arg\min_{\overline{\boldsymbol{\xi}}^*_j} \frac{1}{2}\Vert \overline{\boldsymbol{\xi}}^*_j-\boldsymbol{\xi}^*_j\Vert_2^2 \label{projection_cost}\\
    \textbf{A}\overline{\boldsymbol{\xi}}^*_j = \textbf{b}(\textbf{p}_j), \qquad \textbf{g}(\overline{\boldsymbol{\xi}}^*_j) \leq  \textbf{0} \label{projection_const}
\end{align}
\normalsize

\noindent The cost function \eqref{projection_cost} aims to perform a minimal change to the output of the QP \eqref{lower_cost_reform}-\eqref{lower_eq_reform} in order to satisfy the constraints. The inequalities in \eqref{projection_const} model collision avoidance, kinematic and lane bounds. We present their algebraic form in Appendix \ref{appendix}. Therein, we also show that inequality constraints can be reformulated to induce a special structure in our projection optimization. In particular, \eqref{projection_cost}-\eqref{projection_const} can be reduced to the fixed point iterations \eqref{fixed_point_1}-\eqref{fixed_point_2}, wherein $k$ represents the iteration index.


\small
\begin{align}
    {^{k+1}}\textbf{e}_j, {^{k+1}}\boldsymbol{\lambda}_j = \textbf{h} ({^k} \overline{\boldsymbol{\xi}}_j^*, {^k}\boldsymbol{\lambda}_j ) \label{fixed_point_1}\\
    {^{k+1}}\overline{\boldsymbol{\xi}}^*_j = \arg\min_{\overline{\boldsymbol{\xi}}^*_j} \frac{1}{2}\Vert \overline{\boldsymbol{\xi}}^*_j-\boldsymbol{\xi}^*_j\Vert_2^2 +\frac{\rho}{2} \left\Vert \textbf{F}\overline{\boldsymbol{\xi}}_j^* -{^{k+1}}\textbf{e}_{j} \right\Vert_2^2\nonumber \\-{^{k+1}}\boldsymbol{\lambda}_j^T\overline{\boldsymbol{\xi}}_j^*, \qquad \textbf{A}\overline{\boldsymbol{\xi}}^*_j = \textbf{b}(\textbf{p}_j) \label{fixed_point_2}
\end{align}
\normalsize

\noindent In \eqref{fixed_point_1}-\eqref{fixed_point_2},  $\textbf{F}$ represents a constant matrix and $\textbf{h}$ is some closed-form analytical function. 
We derive these entities in  Appendix \ref{appendix}. The main cost of projection optimization stems from solving the QP \eqref{fixed_point_2}. However, since there are no inequality constraints in \eqref{fixed_point_2}, the QP essentially boils down to an affine transformation of the following form:

\small
\begin{align}
    ({^{k+1}}\overline{\boldsymbol{\xi}}^*_j, {^{k+1}}\nu) = \textbf{M}\boldsymbol{\eta}({^k} \overline{\boldsymbol{\xi}}_j^*),
    \label{affine_trans}
\end{align}
\begin{align}
    \textbf{M} = \begin{bmatrix}
        \textbf{I}+\rho\textbf{F}^T\textbf{F} & \textbf{A}^{T} \\ 
        \textbf{A} & \textbf{0}
    \end{bmatrix}^{-1}, \boldsymbol{\eta} = \begin{bmatrix}
        -\rho\textbf{F}^T {^{k+1}}\textbf{e}_j+{^{k+1}}\boldsymbol{\lambda}_j+\boldsymbol{\xi}_j^*\\
        \textbf{b}(\textbf{p}_j)
    \end{bmatrix} 
\end{align}
\normalsize

Fig.\ref{unrolling} presents an unrolled perspective of our projection optimizer. As can be seen, it takes $\boldsymbol{\xi}^*_j$ as the input along with the initial guess for the solution ${^k}\overline{\boldsymbol{\xi}}_j^*$, and parameter ${^k}\boldsymbol{\lambda}_j$ at $k = 0$. The latter is the so-called Lagrange multiplier associated with inequality constraints. The initial guesses are then gradually updated by recursively passing them through the $\textbf{h}(.)$ and $QP(.)$ blocks a specified number of times. The following important features of our projection optimizer are crucial for building our end-to-end learning pipeline.

\noindent \textbf{Differentiability:} Both the $\textbf{h}(.)$ and $QP(.)$ blocks are differentiable since the former is a closed-form function and the latter reduces to simply an affine transformation \eqref{affine_trans}. This allows us to compute how the output of the projection optimizer will vary if either the input or the initialization values will change. More generally, let $\mathcal{L}(\overline{\boldsymbol{\xi}}^*_j)$ be some loss function defined over the output of the projection optimizer. Due to the differentiability property, we can efficiently obtain the gradients $\nabla_{\boldsymbol{\xi}_j^*}\mathcal{L}$, $\nabla_{{^k}\overline{\boldsymbol{\xi}}_j^*}\mathcal{L}$, $\nabla_{{^k}\boldsymbol{\lambda}_j}\mathcal{L}$, etc.

\noindent \textbf{Batchable Structure:} Besides, being differentiable, we need the projection optimizer to be batchable for it to be easily embedded into the neural network pipeline \cite{amos2017optnet}. In other words, we should be able to compute the projection for several $\boldsymbol{\xi}_j^*$ in parallel. To this end, we recall \eqref{fixed_point_2}-\eqref{affine_trans} and note that the $QP(.)$ block in Fig.\ref{unrolling} essentially reduces to a matrix-vector product that can be trivially batched. Moreover, the matrix \textbf{M} in \eqref{affine_trans} is independent of the batch index and thus needs to be computed only once. In fact, for the learning pipelines discussed later, we pre-store \textbf{M} before the training is started.

\subsection{Supervised Learning with CVAE}

\noindent In this section, we derive a Behaviour Cloning (BC) framework to learn a policy that maps observations $\textbf{o}$ directly to optimal behavioral inputs $\textbf{p}$. Typically in BC, we assume that we have access to a dataset $ (\textbf{o}, \boldsymbol{\tau}_e)$ that demonstrates the expert (optimal) trajectory $\boldsymbol{\tau}_e$ for each observation vector $\textbf{o}$. However, we cannot directly access a demonstration of the optimal behavioral input $\textbf{p}$ employed by the expert. Instead, we have their indirect observation through $\boldsymbol{\tau}_e$. Thus, our problem is more complicated than the typical BC setup. 

\begin{figure}[h!]
\centering
 \includegraphics[scale=0.34]{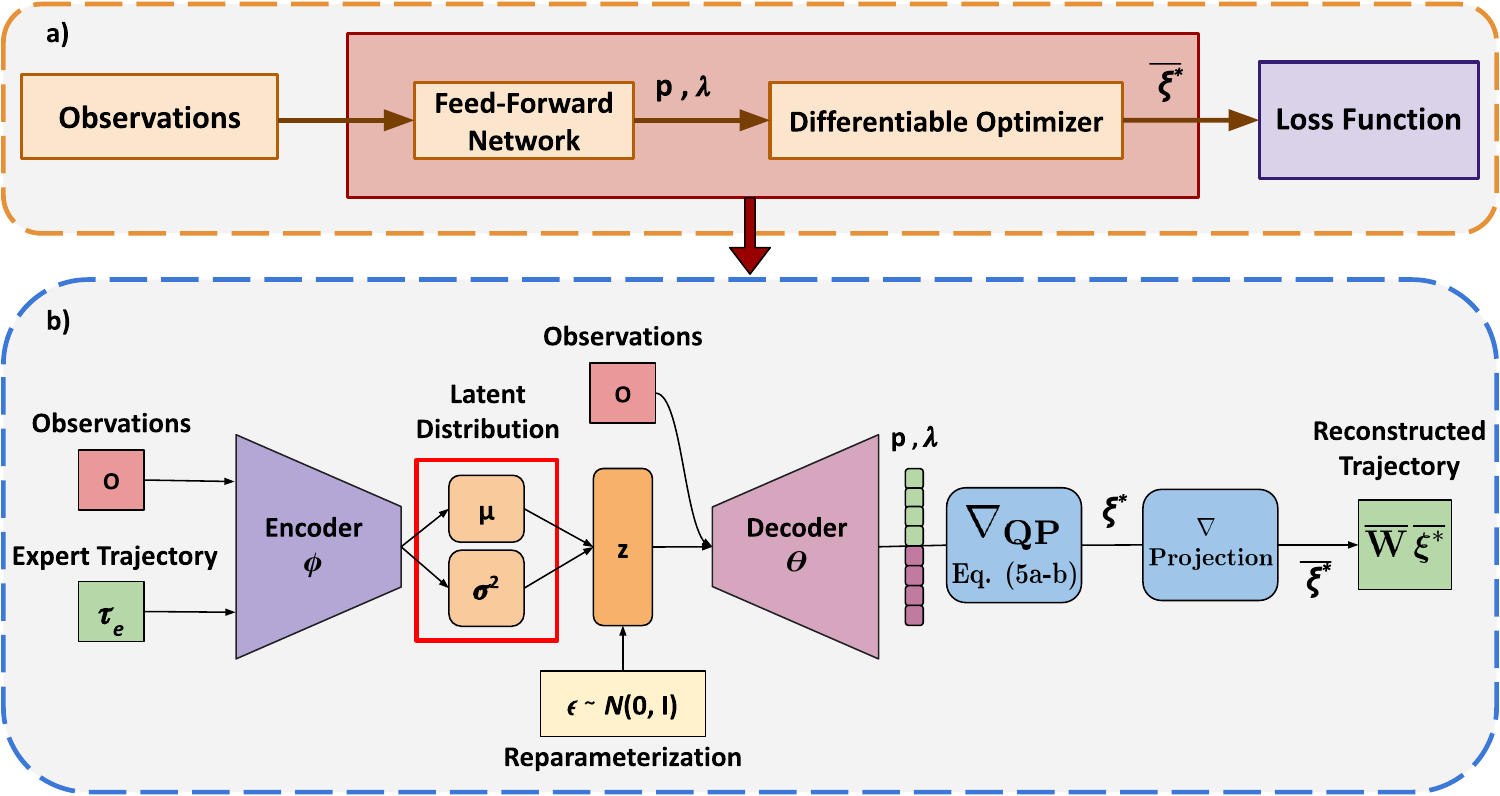}
\caption{(a) Our BC framework features a niche neural network architecture using a combination of feedforward and differentiable optimization layers. Fig.(b) shows the overall CVAE encoder-decoder architecture comprising feedforward layers as MLPs. The differentiable optimizer consists of equality-constrained QP and our custom projection operator.}
\label{cvae_pipeline}
\vspace{-0.2cm}
\end{figure}

We address these challenges using an unconventional architecture combining feedforward and differentiable optimization layers \cite{amos2017optnet} to learn the optimal behavioral inputs from expert trajectory demonstrations. An overview of our approach is illustrated in Fig.\ref{cvae_pipeline} (a). The learnable weights are present only in the feedforward layers. It takes in observations $\textbf{o}$ to output the behavioral inputs $\textbf{p}$ and the Lagrange multipliers $\boldsymbol{\lambda}$ (recall \eqref{fixed_point_2}), which is fed to the differentiable optimizer resulting in optimal trajectory coefficients $\overline{\boldsymbol{\xi}}^*$. The BC loss is computed over $\overline{\boldsymbol{\xi}}^*$. The backpropagation required for updating the weights of the feedforward layer needs to trace the gradient of the loss function through the optimization layer. 


\noindent \textbf{Need for CVAE:} We want our learned policy to induce a distribution over $\textbf{p}$ so that for each observation $\textbf{o}$, we can then draw samples $\textbf{p}_j$ from it and solve the trajectory optimizations conditioned on them. With this motivation, we use a deep generative model called CVAE \cite{NIPS2015_8d55a249}, illustrated in Fig.\ref{cvae_pipeline} (b) as our learning pipeline. It consists of an encoder-decoder architecture constructed from a multi-layer perceptron (MLP) with weights $\boldsymbol{\phi}$ and $\boldsymbol{\theta}$ respectively. The decoder network also has an optimization layer that takes the output ($\textbf{p}$) of its MLP to produce an estimate of optimal trajectory coefficients  $\overline{\boldsymbol{\xi}}^*$. 



The encoder network maps $(\textbf{o}, \boldsymbol{\tau}_e)$ to a latent variable $\textbf{z}$ with distribution $\textbf{q}_{\boldsymbol{\phi}}(\boldsymbol{\mu}(\boldsymbol{\phi}), \boldsymbol{\Sigma}(\boldsymbol{\phi}) )$. The covariance matrix $\boldsymbol{\Sigma}(\boldsymbol{\phi})$ is diagonal and formed with the vector $\boldsymbol{\sigma}^2$ produced by the encoder. The decoder then maps this latent distribution to $\textbf{p}_{\boldsymbol{\theta}}(\overline{\boldsymbol{\xi}}^*|\textbf{z}, \textbf{o})$ through its MLP and optimization layers. In the training (offline) phase, both the networks are trained end-to-end with loss function \eqref{eq_cvae}, where $\overline{\textbf{W}} = \begin{bmatrix}
    \textbf{W} & \textbf{0}\\
    \textbf{0} & \textbf{W}
\end{bmatrix}$. The first term is the reconstruction loss responsible for bringing the output of the decoder network as close as possible to the expert trajectory. The second term in \eqref{eq_cvae} acts as a regularizer that aims to make the learned latent distribution $\textbf{q}_{\boldsymbol{\phi}}(\textbf{z} | \textbf{o}, \boldsymbol{\tau}_e)$ as close as possible to the prior isotropic normal distribution $\mathcal{N}(\textbf{0}, \textbf{I})$. The $\boldsymbol{\beta}$ hyper-parameter acts as a trade-off between the two cost terms. 

\vspace{-0.3cm}
\small
\begin{equation}
    \mathcal{L}_{\textit{CVAE}} = \sum \Vert \overline{\textbf{W}} \, \overline{\boldsymbol{\xi}}^*(\boldsymbol{\theta}, \boldsymbol{\phi}) - \boldsymbol{\tau}_e \Vert_2^2 \newline
    + \beta \, {D}_{\mathbf{KL}}[\textbf{q}_{\boldsymbol{\phi}}(\textbf{z}\,| \,\textbf{o}, \,\boldsymbol{\tau}_e)\, | \mathcal{N}(\textbf{0}, \textbf{I})]  \label{eq_cvae}
\end{equation}
\normalsize
\vspace{-0.3cm}

In the inferencing (online) phase, we draw samples of $\textbf{z}$ from the prior isotropic normal distribution and then pass them through the decoder MLP to get samples of optimal behavioural inputs $\textbf{p}$ along with $\boldsymbol{\lambda}$. Finally, these are passed through the optimization layers to generate distribution for the optimal trajectory coefficients $\overline{\boldsymbol{\xi}}^*$.

\noindent\textbf{Incorporating Self Supervision Loss:} Let us assume a simplified world model where the neighboring vehicles are non-reactive dynamic obstacles. Moreover, we have some approximate predictions for their trajectories over a future time horizon. For example, we can take the current velocity and positions of the neighboring vehicles from the observation vector $\textbf{o}$ and perform a linear prediction. Under this simplified world mode, we can augment the meta-cost \eqref{meta_cost} into our learning pipeline as a self-supervision cost. That is, we can modify our loss function as 

\vspace{-0.3cm}
\small
\begin{align}
    \mathcal{L}_{cvae}+s \sum_{k}c_{m}({^k}\overline{\boldsymbol{\xi}}^*),
\end{align}
\normalsize
\noindent where ${{^k}}\overline{\boldsymbol{\xi}}^*$ is the output at the $k^{th}$ stage (iteration) of the unrolling. The scalar $s$ trades off the CVAE and self-supervision loss. As discussed in \eqref{lower_cost_reform}, a part of $(c_m)$ is the constraint (kinematic, collision, lane) residuals which are in fact same as the residuals of our projection optimizer. Thus, the addition of the self-supervision loss forces the network to learn $\textbf{p}$ such that it aids in faster convergence of the optimizer.


\newtheorem{remark}{Remark}\label{rem_1}
\begin{remark}
    We used the output of the QP \eqref{lower_cost_reform}-\eqref{lower_eq_reform} in Fig.\ref{cvae_pipeline} along with the predicted $\boldsymbol{\lambda}$ from the decoder MLP to initialize the differentiable projection layer. This design choice couples the learning of behavioral inputs and the initialization for the lower-level trajectory optimizer.
\end{remark}

\subsection{Self-Supervised Learning with MLP}
\begin{figure}[h!]
\centering
 \includegraphics[scale=0.36]{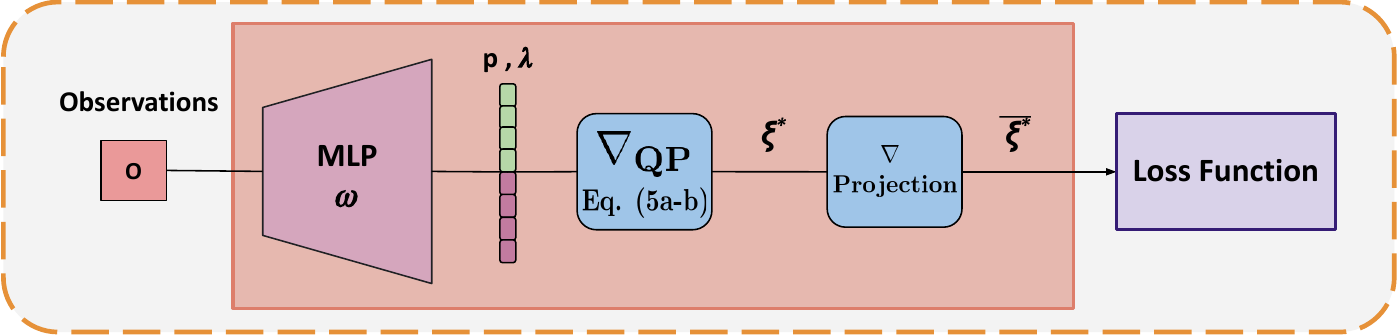}
\caption{Fig. shows an MLP combined with differentiable optimization layers used for self-supervised learning of behavioural inputs}
\label{mlp_pipeline}
\vspace{-0.5cm}
\end{figure}

\noindent In this subsection, we formulate a behavioral input learning pipeline with purely self-supervision loss. The primary motivation stems from the fact that demonstrations could be sub-optimal or hard to obtain in dense traffic conditions. To this end, we construct a simple feedforward network using an MLP (see Fig.\ref{mlp_pipeline} ) with learnable parameter $\boldsymbol{\omega}$. The network is trained with the loss function \eqref{ss_loss}.

\vspace{-0.5cm}
\small
\begin{align}
    \underset{\boldsymbol{\omega}}{min.} \, \,  \mathbb{E}_{\textbf{o} \sim p_{\textbf{(o)}}} c_m(\boldsymbol{\pi}_{\boldsymbol{\omega}}(\textbf{o}); \textbf{o}) \approx \frac{1}{n}\sum_jc_m(\boldsymbol{\pi}_{\boldsymbol{\omega}}(\textbf{o}_j); \textbf{o}_j)
    \label{ss_loss}
\end{align}
\normalsize

\noindent where $\boldsymbol{\pi}_{\boldsymbol{\omega}} (.)$ represents the MLP policy of Fig.\ref{mlp_pipeline} that take in observations $\textbf{o}$ and outputs the optimal trajectory coefficients. The Expectation operator in \eqref{ss_loss} is approximated by empirical mean. The observation samples are the ones encountered during the collection of expert demonstrations for supervised learning. 

The learned MLP provides only a single output. However, our planning approach needs a distribution from where multiple samples of $\textbf{p}, \boldsymbol{\lambda}$ can be drawn. Thus, we treat the output of the MLP as the mean of a Gaussian distribution. For the Covariance, we use a constant diagonal matrix.

\section{Connections to Existing Works} \label{connections}
\subsubsection*{Trajectory Sampling Approaches} As mentioned earlier, \cite{wei2014behavioral}, \cite{fernet_planner} sample behavioral inputs from a pre-discretized grid that is oblivious to how the resulting trajectories are performing on the driving task. 
Authors in \cite{sun2022fiss} address this drawback to some extent as they adapt the sampling strategy based on optimal trajectories obtained in the past planning cycles. However, none of these cited works explicitly enforce collision avoidance constraints in their approach. Our prior work \cite{adajania2022multi} addressed constraint handling but the behavioral input sampling was still handcrafted. 


\subsubsection*{Differentiable Optimization Layers} Embedding optimization layers into neural network pipelines has recently garnered much attention. Although technically, any off-the-shelf optimizer can be embedded into neural architectures \cite{gould2021deep}, the efficiency of the resulting training could be limited. Thus, a strong focus has been on developing batchable GPU accelerated optimizers \cite{amos2017optnet}. Our projection optimizer satisfies both of these requirements. Moreover, its unique structure allows us to avoid matrix factorizations during training (recall \eqref{affine_trans}) as these can be pre-stored. As a result, our whole training pipeline ran stably on 32-bit precision on Graphical Processing Units (GPU)s. In contrast, \cite{amos2017optnet} strongly recommends running their differentiable optimizer in 64-bit, which could be slow.

\section{Validation and benchmarking}
In this section, we qualitatively validate the performance of our projection optimizer and answer the following research questions:

\begin{itemize}
    \item \textbf{Q1:} How do learned behavioral inputs perform as compared to handcrafted heuristics?
    \item \textbf{Q2:} How does our approach fare compare to the State-of-the-art trajectory planners and Model Predictive Control (MPC) methods?
\end{itemize}

\begin{figure}[h!]
\centering
 \includegraphics[scale=0.34]{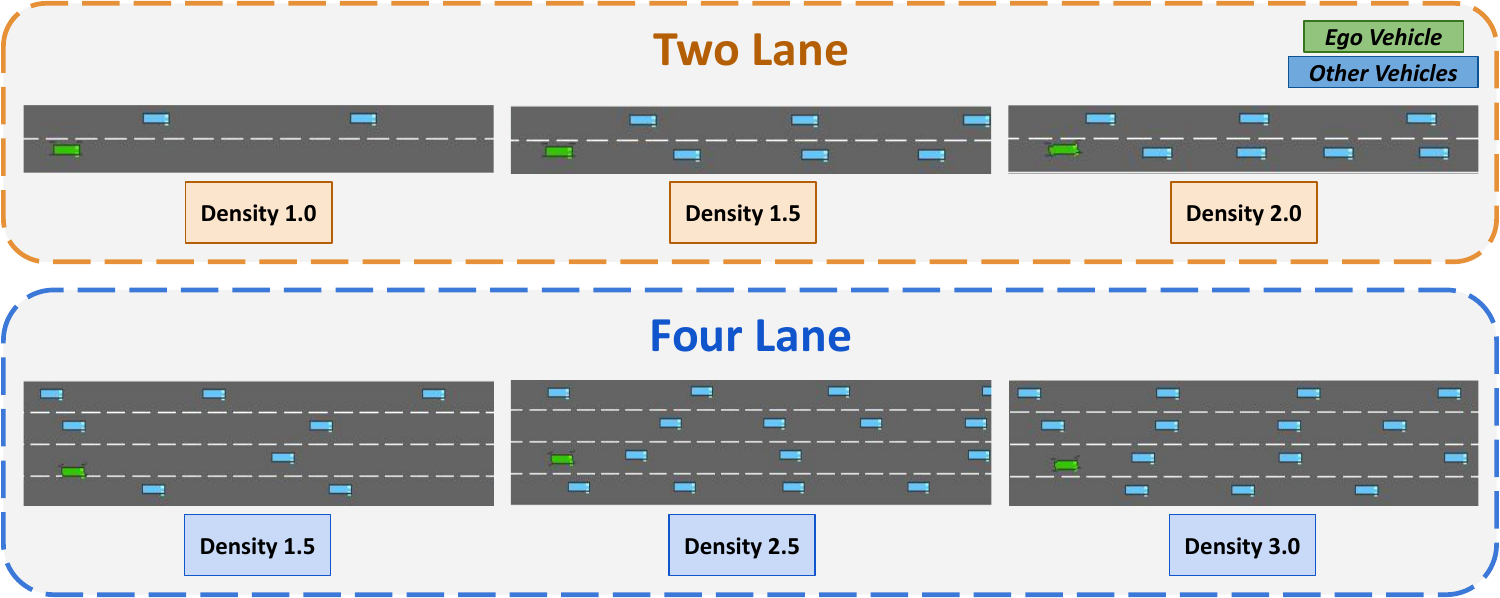}
\caption{Two and four-lane highway driving scenarios with varying traffic density used for benchmarking our approach with different baseline MPCs.}
\label{benchmark}
\vspace{-0.5cm}
\end{figure}

\subsection{Implementation Details}
\noindent We implemented our trajectory planner comprising of QP \eqref{lower_cost_reform}-\eqref{lower_eq_reform} and projection \eqref{projection_cost}-\eqref{projection_const} in Python using JAX \cite{jax} library as our GPU-accelerated linear algebra back-end. The matrix $\textbf{W}$ in \eqref{param} is constructed from a $10^{th}$ order polynomial. We also created equivalent PyTorch implementations for embedding into the training pipeline. Our simulation pipeline was built on the Highway Environment (highway-env) simulator \cite{Leurent_An_Environment_for_2018}. The neighboring vehicles use simple rule-based approach for lateral and longitudinal control.

\subsubsection{Hyper-parameter Selection} The behavioral input $\textbf{p}$ was modeled as four set-points for lateral offsets and desired longitudinal velocities. That is, $\textbf{p} = \begin{bmatrix} y_{d, 1}, \dots, y_{d,4}, v_{d, 1}, \dots, v_{d,4}   \end{bmatrix}$. We divided the planning horizon into four segments and associated one pair of lateral offsets and desired velocity to each of these. 

\subsubsection{CVAE and MLP Training} The details of the encoder-decoder network architecture of our CVAE are presented in the accompanying video. During training, the input to the CVAE is the expert trajectory and a 55-dimensional observation vector ($\textbf{o}$), containing the state of the ego-vehicle, the ten closest obstacles, and the road boundary. For the ego-vehicle, the state consists of heading, lateral and longitudinal speeds. The obstacle state consists of longitudinal/lateral positions and the corresponding velocities for the ten closest obstacles. We express all the position-level information with respect to the center of the ego vehicle. During inference, the decoder network of CVAE only needs $\textbf{o}$, and samples $\textbf{z}$ are drawn from an isotropic Gaussian. For MLP, only the observation vector is needed.

We used the cross-entropy method \cite{botev2013cross}, run offline with a batch size of 5000, to collect the demonstration of optimal trajectories for training our CVAE. We note that our demonstrations could be sub-optimal and sparse.  However, even with such a simple data set, our CVAE and MLP were able to learn valuable behavioral inputs. 


\subsubsection{Baselines} We used our trajectory planner in a receding horizon manner to create two MPC variants. We will henceforth refer to it as MPC-Supervised and MPC-Self-Supervised depending on whether the behavioral inputs are obtained from either supervised CVAE or self-supervised MLP. Both the MPC variants take the same observation vector $\textbf{o}$ as the input and output coefficients of the optimal trajectories. These are converted to steering and acceleration input vectors. We compare our MPC with the following baselines and SOTA approaches:


\noindent \textbf{MPC-Grid}: This baseline operates with handcrafted behavioral inputs. The vector $\textbf{p}$ consists of some set-points for lateral offsets and desired velocities sampled from a pre-specified grid instead of a neural network. The grid is centered around the lane center-line and desired speed.\\
\noindent \textbf{Batch-MPC} of \cite{adajania2022multi}: This SOTA MPC uses a different set of behavioral inputs, namely goal positions for the longitudinal and lateral components of the trajectory. That is, $\textbf{p} = \begin{bmatrix} x_f, y_f  \end{bmatrix}$. Again, the behavioral input is sampled from a pre-specified grid.

\noindent \textbf{Model Predictive Path Integral (MPPI)} \cite{williams2017model}: This is the SOTA approach for receding horizon planning. It operates by sampling trajectories, evaluating the cost $c_m$ (recall \eqref{meta_cost}) and then updating the sampling distribution. The MPPI baseline directly works in the space of trajectories and does not use any behavioral input. We leverage the insight presented in \cite{bhardwaj2022storm} where the covariance matrix is also adapted for better optimization.

\begin{remark} \label{des_dirac}
MPC-Grid has the same trajectory planner as our MPC-Supervised and MPC-Self-supervised variants. The only difference stems from the sampling of behavioral inputs. Batch-MPC \cite{adajania2022multi} also explicitly enforces kinematic and collision avoidance constraints. In contrast, MPPI operates by rolling all the constraints as penalties in the cost function. 
\end{remark}

\subsubsection{Environments, Tasks, and Metrics}
\noindent The highway driving scenarios are presented in Fig.\ref{benchmark}. For each scenario, we had three different traffic densities. We use the internal parameter of highway-env named "density" to control how closely each vehicle is placed at the start of the simulation. We evaluate two sets of 50 configurations spawned using different random seeds for each density in two or four-lane driving settings.  We fixed the random seed of the simulator to ensure that all MPC baselines are tested across the same set of traffic configurations.


The task in the experiment was for the ego-vehicle to drive as fast as possible without colliding with the obstacles and going outside the lane boundary. Thus, $v_{des} = v_{max}$ was used in the meta-cost \eqref{meta_cost}. Our evaluation metric has two components: (i) collision rate and (ii) average velocity achieved within an episode. Since the ego-vehicle can achieve arbitrary high velocity while driving rashly, we only consider velocities from collision-free episodes.

\vspace{-0.1cm}
\subsection{Empirical Validation of Projection Optimizer}
\begin{figure}[h!]
\centering
 \includegraphics[scale=0.30]{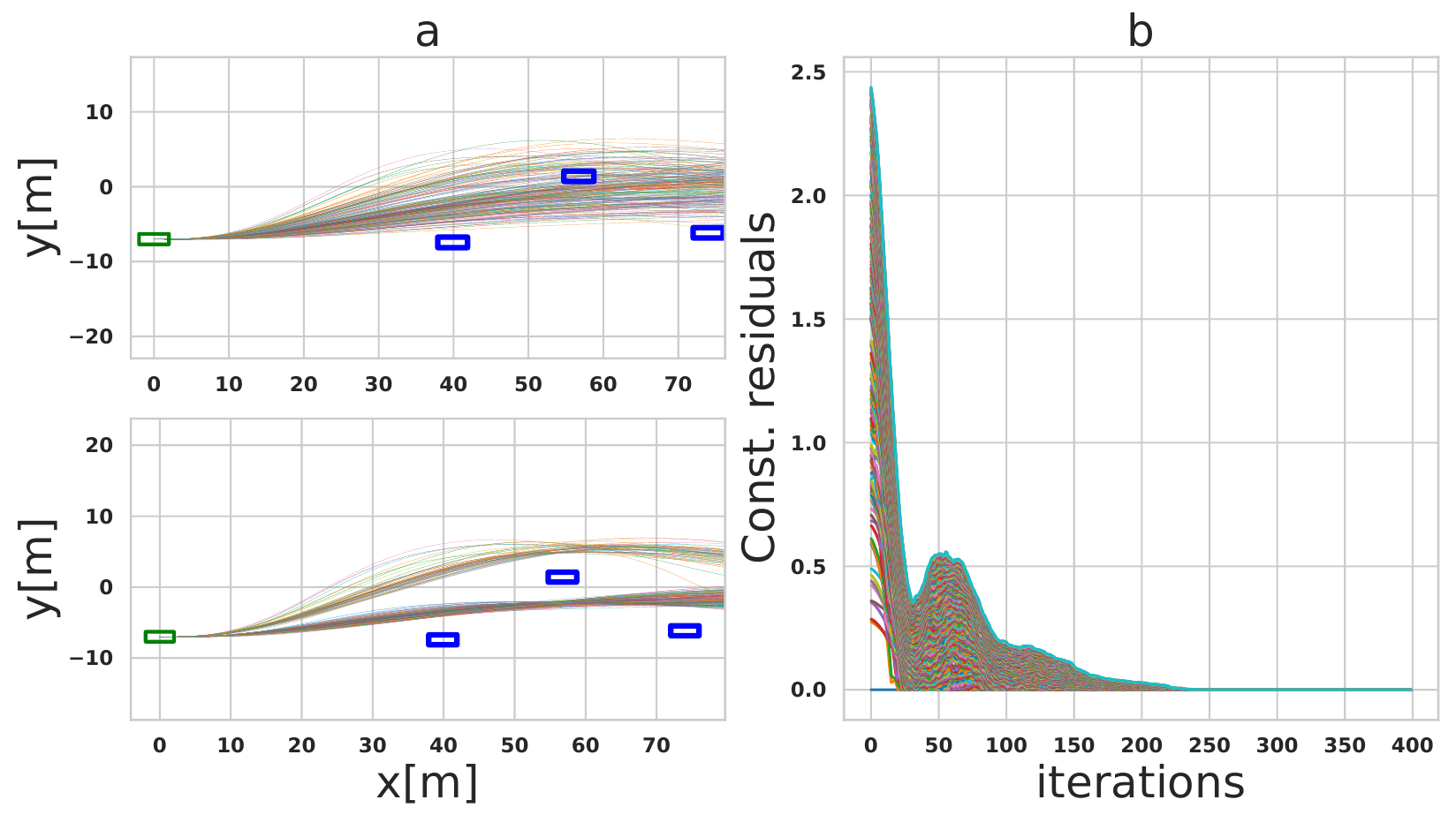}
\caption{Fig.(a) (top): Trajectories produced by QP \eqref{lower_cost_reform}-\eqref{lower_eq_reform} for randomly sampled $\textbf{p}_j$. The green and blue rectangle represents the ego vehicle and the obstacle, respectively. Fig.(a)(bottom): The projection of trajectories onto the feasible set of collision avoidance and velocity/acceleration bounds. Fig.(b): The trend of constraint residuals across iterations for every instance in the batch. Most trajectories residuals converge to zero within 100 iterations.}
\label{projection_qual}
\vspace{-0.5cm}
\end{figure}

\noindent Fig.\ref{projection_qual}(a) shows a typical output of our projection optimizer for a scene with static obstacles (blue rectangles). We consider 400 randomly sampled $\textbf{p}_j$ that were passed to QP \eqref{lower_cost_reform}-\eqref{lower_eq_reform} resulting in trajectory distribution shown in Fig.\ref{projection_qual}(a)(top). These were then passed to our projection optimizer \eqref{projection_cost}-\eqref{projection_const} that led to collision-free trajectories residing in different homotopies. Fig.\ref{projection_qual}(b) shows the constraint residuals across iterations for every instance of the batch. Typically, 100 iterations were enough to drive the constraint residuals to zero for a majority of the trajectory samples. 

\subsection{Effect of Learned Behavioural Inputs}
\begin{figure}[h!]
\centering
 \includegraphics[scale=0.33]{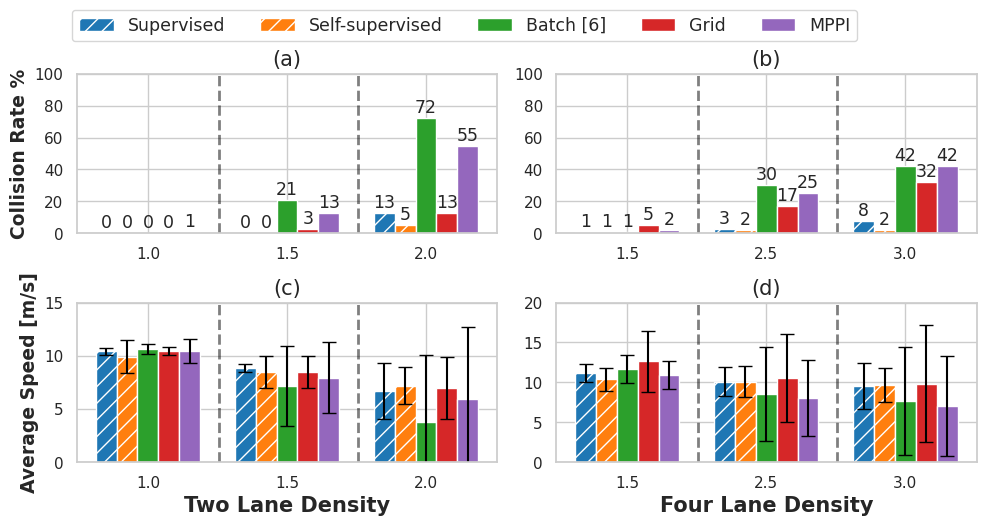}
\caption{Comparison of MPC baselines with ours; MPC-Supervised and MPC-Self-supervised in two-lane and four-lane driving scenarios}
\label{benchmark_comparison}
\vspace{-0.4cm}
\end{figure}

\noindent A two-lane driving scenario offers minimal scope for maneuvers. Thus, in a low-traffic density, all the baselines and our two approaches perform equally well (Fig.\ref{benchmark_comparison}(a)). This shows that a simple handcrafted grid search performed in MPC-Grid and Batch-MPC of \cite{adajania2022multi} is enough to come up with the right set of behavioral inputs. Moreover, the performance of MPPI shows that one can even bypass the behavioral input sampling altogether and search directly in the space of trajectories. 


As the traffic density increases in the two-lane scenarios, we can see the benefit of behavioral input sampling (MPC-Grid outperforming MPPI) and, even more importantly, going beyond the handcrafted heuristics (ours outperforming hand-crafted behavioral inputs sampling). The trend is particularly stark in dense four-lane scenarios where our MPC-Supervised and MPC-Self-supervised provide a $4-10\times$ reduction in collision rate (Fig.\ref{benchmark_comparison}(b)). Fig.\ref{benchmark_comparison}(c)-(d) show that the average speed achieved by our approaches is either better or with the baselines in all traffic densities.

Among our approaches, MPC-Supervised performs better in medium traffic densities; two-lane(1.0, 1.5) and four-lane (1.5, 2.5). In contrast, MPC-Self-supervised outperforms in two and four-lane scenarios with a traffic density of 3.0. This can be attributed to the fact that the expert demonstrations were sparse in challenging scenarios. Moreover, this pattern also showcases the importance of our self-supervised learning pipeline.

Table \ref{table_comp_time} correlates the number of iterations of our projection optimizer with collision rate and max achieved speed. As can be seen, the learned behavioral inputs also aid in the convergence of the optimizer. For those learned from the supervised training, the projection optimizer needs around 75 iterations to achieve its best performance, documented in Fig.\ref{benchmark_comparison}. The self-supervised training led to even faster convergence. 

\begin{remark} \label{rem_mppi}
The poor performance of MPPI is attributed to two reasons. First, we have observed that sampling in the space of behavioral inputs provides a more focused search than sampling direct trajectories for autonomous driving benchmarks. Second, MPPI rolls the collision constraints into the costs, and thus this soft-constraint handling proves detrimental in dense scenarios.
\end{remark}

\begin{remark} \label{rem_error_bar}
The success rate in Fig.\ref{benchmark_comparison}(a)-(b) is based on the number of collision-free runs across all the episodes for a particular benchmark. Hence, this metric doesn't have an error bar. In contrast, the velocity profiles of Fig.\ref{benchmark_comparison} vary within a simulation episode and across the whole data set. Thus, we present the error bars to capture this variability.
\end{remark}

\vspace{-0.3cm}
\begin{table}[!h]
\centering
\scriptsize
\caption{Max. Iteration vs Collision rate}
\begin{tblr}{
  row{1} = {c},
  row{4} = {c},
  cell{1}{1} = {c=4}{},
  cell{2}{2} = {c=3}{c},
  cell{3}{2} = {c},
  cell{3}{3} = {c},
  cell{3}{4} = {c},
  cell{5}{1} = {c},
  cell{6}{1} = {c},
  cell{6}{2} = {c},
  vlines,
  hline{1-2,4-7} = {-}{},
  hline{3} = {2-4}{},
}
\textbf{Average Collision Rate \% / Speed [m/s]} &                                           &              &          \\
                                                 & \textbf{\textit{Projection iteration}} &              &          \\
                                                 & 25                                        & 50           & 75       \\
Supervised                                       & 20 / 8.15                                 & 14 / 8.95    & 8 / 9.59 \\
Self-Supervised                                  & \, 2 / 9.545                               & \, 2 / 9.69  & \, 2 / 9.65  \\
Grid                                             & 31 / 10.42                                & 36 / 9.41           & 31 / 10.04          
\end{tblr}
\label{table_comp_time}
\normalsize
\end{table}

\subsection{Ablation: Effect of Training with Projection Layer}

\begin{figure}[h!]
\centering
 \includegraphics[scale=0.30]{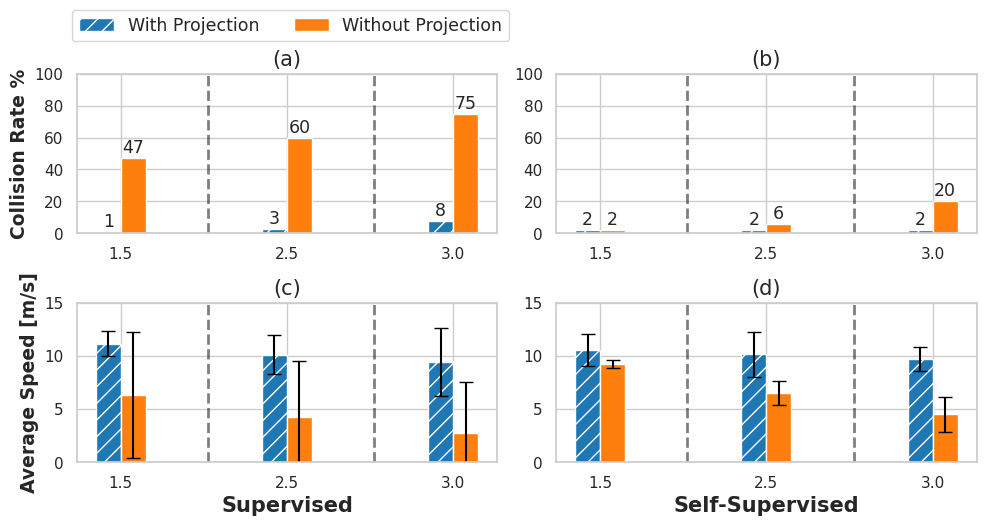}
\caption{Driving performance achieved with neural networks trained with and without our differentiable projection optimizer.}
\label{ablation}
\vspace{-0.6cm}
\end{figure}

\noindent Fig.\ref{ablation} presents the second key result of our work. It showcases the importance of embedding our custom projection operator in the training pipeline shown in Fig.\ref{cvae_pipeline} and Fig.\ref{mlp_pipeline}. As can be seen, the performance severely degrades in the absence of the projection layer because the network does not get corrective feedback from the optimizer during training. An alternative to our approach could be to directly penalize the network output. However, as shown in \cite{pulver2021pilot}, such an approach shows poor generalization.


We note that our supervised approach shows higher degradation in performance than the self-supervised variant. We believe this is due to the neural network predictions mimicking the (sub-optimal) expert demonstration at the cost of violating the constraints. For self-supervised training, such conflicting objectives do not exist.


\section{Conclusions and Future Work}
We showed how behavioral inputs can be learned while considering the ability of the downstream trajectory optimizer. To this end, we proposed a differentiable optimizer and embedded it as a layer in a neural network. We adopted both supervised and self-supervised learning approaches. The latter generalized better in high-density traffic scenarios where expert demonstrations are hard to obtain and thus sparse. To validate our approach, we extensively compared against strong baselines, including MPPI and \cite{adajania2022multi}. 
Finally, we showed how training without our projection optimizer leads to severely degraded performance due to the lack of constraints on the neural network predictions.

Our differentiable optimizer opens-up new possibilities, especially in the context of autonomous navigation. The specialized structure offers computational advantages over the off-the-shelf libraries like \cite{amos2017optnet} designed for a broader application spectrum.
\vspace{-0.2cm}
\section{Appendix}\label{appendix}


\begin{table*}[!t]
\centering
\caption{\scriptsize{List of Inequality Constraints Used in the projection optimization}}
\small 
\begin{tabular}{|c|c|c|c|c|c|}
\hline
Constraint Type & Expression & Parameters   \\ \hline
Collision Avoidance  & $-\frac{(x(t)-x_{o, i}(t))^2}{a^2}-\frac{(y(t)-y_{o, i}(t))^2}{b^2}+1\leq 0$ & \makecell{$\frac{a}{2}, \frac{b}{2}$: axis of the circumscribing ellipse \\ of vehicle footprint. \\ $x_{o,i}(t), y_{o, i}(t)$: trajectory of neighboring vehicles} \\ \hline
Velocity bounds & $\sqrt{\dot{x}(t)^2+\dot{y}(t)^2}\leq v_{max}$ & $v_{max}$: maximum velocity of the ego-vehicle   \\ \hline
Acceleration bounds & $\sqrt{\ddot{x}(t)^2+\ddot{y}(t)^2}\leq a_{max}$ & $a_{max}$: maximum acceleration of the ego-vehicle  \\ \hline
Lane boundary & $l_{lb}\leq y(t)\leq l_{ub}$ & \makecell{$y_{lb}, y_{ub}$: Lane bounds.} \\ \hline
\end{tabular}
\normalsize
\label{ineq_list}
\vspace{-0.6cm}
\end{table*}
\normalsize

\noindent \textbf{Reformulating Constraints:} Table \ref{ineq_list} presents the list of all the constraints included in our projection optimizer. The collision avoidance constraints presented there can be re-written in the following form:

\small
\begin{align}
    \textbf{f}_{o, i} = \left \{ \begin{array}{lcr}
x(t) -x_{o, i}(t)-d_{o, i}(t)\cos\alpha_{o, i}(t) \\
y(t) -y_{o, i}(t)-d_{o, i}(t)\sin\alpha_{o, i}(t) \\ 
\end{array} \right \} d_{o, i}(t)\geq 1
\label{sphere_proposed}
\end{align}
\normalsize
\vspace{-0.1cm}
\noindent where $\alpha_{o, i}(t)$ represents the angle that the line-of-sight vector between the ego-vehicle and its $i^{th}$ neighbor makes with the $X$ axis. Similarly, the variable $d_{o, i}(t)$ represents the ratio of the length of this vector with the minimum distance separation required for collision avoidance. Following a similar approach, we can rephrase the velocity and acceleration bounds from Table \ref{ineq_list} as:

\vspace{-0.3cm}

\small
\begin{align}
    \textbf{f}_{v} = \left \{ \begin{array}{lcr}
\dot{x}(t) -d_{v}(t)\cos\alpha_{v}(t) \\
\dot{y}(t) -d_{v}(t)\sin\alpha_{v}(t)\\ 
\end{array} \right \}, v_{min}\leq d_{v}(t)\leq v_{max}
\label{vel_bound_proposed}
\end{align}
\normalsize

\vspace{-0.5cm}
\small
\begin{align}
    \textbf{f}_{a} = \left \{ \begin{array}{lcr}
\ddot{x}(t) -d_{a}(t)\cos\alpha_{a}(t) \\
\ddot{y}(t) -d_{a}(t)\sin\alpha_{a}(t)\\ 
\end{array} \right \}, 0\leq d_{a}(t)\leq a_{max}
\label{acc_bound_proposed}
\end{align}
\normalsize

The variables $\alpha_{o, i}(t)$, $\alpha_{o, i}(t)$, $\alpha_{a, i}(t)$, $d_{o, i}(t)$, $d_{v, i}(t)$, and $d_{a, i}(t)$  are additional variables that our batch projection optimizer will obtain along with $\overline{\boldsymbol{\xi}}_j^*$.

\noindent \textbf{Reformulated Problem:} Using the developments in the previous section and the trajectory parametrization presented in \eqref{param}, we can now replace the projection optimization \eqref{projection_cost}-\eqref{projection_const} with the following. Note that \eqref{lane_reform} is the matrix representation of the lane boundary constraints presented in Table \ref{ineq_list}.

\vspace{-0.8cm}
\small
\begin{subequations}
\begin{align}
    \overline{\boldsymbol{\xi}}_j^{*} = \arg\min_{\overline{\boldsymbol{\xi}}^*_j}\frac{1}{2}\Vert \overline{\boldsymbol{\xi}}^*_j-{\boldsymbol{\xi}}_j^*\Vert_2^2\label{cost_reform}  \\
    \textbf{A} \overline{\boldsymbol{\xi}}^*_j= \textbf{b}(\textbf{p}_j) \label{eq_reform} \\
    \widetilde{\textbf{F}} \hspace{0.05cm} \overline{\boldsymbol{\xi}}^*_j = \widetilde{\textbf{e}}(\boldsymbol{\alpha}_j, \textbf{d}_j) \label{nonconvex_reform}  \\
    \textbf{d}_{min} \leq \textbf{d}_j\leq \textbf{d}_{max} \label{d_reform_1}\\
     \textbf{G}\overline{\boldsymbol{\xi}}^*_j \leq \textbf{y}_{lane} \label{lane_reform}
\end{align}
\end{subequations}
\normalsize
\vspace{-0.5cm}
\small
\begin{align}
    \widetilde{\textbf{F}} = \begin{bmatrix}
    \begin{bmatrix}
    \textbf{F}_{o}\\
    \dot{\textbf{W}}\\
    \ddot{\textbf{W}}
    \end{bmatrix} & \textbf{0}\\
    \textbf{0} & \begin{bmatrix}
    \textbf{F}_{o}\\
    \dot{\textbf{W}}\\
    \ddot{\textbf{W}}
    \end{bmatrix} 
    \end{bmatrix}, \widetilde{\textbf{e}} = \begin{bmatrix}
    \textbf{x}_o+a \textbf{d}_{o, j}\cos\boldsymbol{\alpha}_{o, j}\\
     \textbf{d}_{v, j}\cos\boldsymbol{\alpha}_{v, j}\\
  \textbf{d}_{a, j}\cos\boldsymbol{\alpha}_{a, j}\\
 \textbf{y}_o+a \textbf{d}_{o, j}\sin\boldsymbol{\alpha}_{o, j}\\
     \textbf{d}_{v, j}\sin\boldsymbol{\alpha}_{v, j}\\
  \textbf{d}_{a, j}\sin\boldsymbol{\alpha}_{a, j}\\
    \end{bmatrix},
\end{align}
\vspace{-0.3cm}
\begin{align}
    \textbf{G} = \begin{bmatrix}
        \textbf{W}\\
        -\textbf{W}
    \end{bmatrix}, \textbf{y}_{lane} = \begin{bmatrix}
        y_{ub} & \dots y_{ub} & y_{lb} \dots y_{lb}
    \end{bmatrix}^T
\end{align}
\vspace{-0.3cm}
\begin{align*}
    \boldsymbol{\alpha}_j = (\boldsymbol{\alpha}_{o, j}, \boldsymbol{\alpha}_{a,j}, \boldsymbol{\alpha}_{v,j}), \qquad \textbf{d}_j =  (\textbf{d}_{o, j}, \textbf{d}_{v, j}, \textbf{d}_{a, j})
\end{align*}
\normalsize

\noindent Constraints \eqref{nonconvex_reform}-\eqref{lane_reform} acts as substitutes for $\textbf{g}(\overline{\boldsymbol{\xi}}_j^*)\leq 0 $ in the projection optimization \eqref{projection_cost}-\ref{projection_const}. 

The matrix $\textbf{F}_o$ is obtained by stacking the matrix $\textbf{W}$ from (\ref{param}) as many times as the number of neighboring vehicles considered for collision avoidance at a given planning cycle. The vector $\textbf{x}_o, \textbf{y}_o$ is formed by appropriately stacking $x_{o, i}(t), y_{o, i}(t)$ at different time instants and for all the neighbors. Similar construction is followed to obtain $\boldsymbol{\alpha}_{o}, \boldsymbol{\alpha}_{v}, \boldsymbol{\alpha}_{a}, \textbf{d}_{o}, \boldsymbol{d}_{v}, \boldsymbol{d}_{a}$. The vector $\textbf{y}_{lane}$ is formed by stacking the upper and lower lane bounds after repeating them $m$ times (planning horizon). Similarly,  vectors $d_{min}, d_{max}$ are formed by stacking the lower and upper bounds for $d_{o, i}(t), d_a(t), d_v(t)$. Note that the upper bound for $d_{o, i}(t)$ can be simply some large number (recall \eqref{sphere_proposed}). Moreover, these bounds are the same across all batches.

\noindent \textbf{Solution Process:} We relax the non-convex equality \eqref{nonconvex_reform} and affine inequality constraints as $l_2$ penalties and augment them into the projection cost \eqref{cost_reform}.

\small
\begin{dmath}
    \mathcal{L} = \frac{1}{2}\left\Vert \overline{\boldsymbol{\xi}}^*_j-\boldsymbol{\xi}^*_j\right\Vert_2^2- \boldsymbol{\lambda}_{j}^T \overline{\boldsymbol{\xi}}^*_j+\frac{\rho}{2} \left \Vert \widetilde{\textbf{F}} \overline{\boldsymbol{\xi}}^*_j-\widetilde{\textbf{e}}\right \Vert_2^2+  \frac{\rho}{2}\left \Vert \mathbf{G} \overline{\boldsymbol{\xi}}^*_{j} - \textbf{y}_{lane} + \mathbf{s}_j \right \Vert^2 = \frac{1}{2}\left\Vert \overline{\boldsymbol{\xi}}^*_j-\boldsymbol{\xi}^*_j\right\Vert_2^2-\boldsymbol{\lambda}_{j}^T \overline{\boldsymbol{\xi}}^*_j+\frac{\rho}{2} \left \Vert \textbf{F} \overline{\boldsymbol{\xi}}^*_j-\textbf{e}\right \Vert_2^2
    \label{aug_lag}
\end{dmath}
\normalsize

\small
\begin{align}
    \textbf{F} = \begin{bmatrix}
        \widetilde{\textbf{F}}\\
        \textbf{G}
    \end{bmatrix}, \textbf{e} = \begin{bmatrix}
        \widetilde{\textbf{e}}\\
        \textbf{y}_{lane}-\textbf{s}_j
    \end{bmatrix}
\end{align}
\normalsize
\noindent Note, the introduction of the Lagrange multiplier $\boldsymbol{\lambda}$ that drives the residual of the second and third quadratic penalties to zero. We minimize \eqref{aug_lag} subject to \eqref{eq_reform} through Alternating Minimization (AM), which reduces to the following steps \cite{masnavi2022visibility}.

\vspace{-0.6cm}
\small
\begin{subequations}
    \begin{align}
        {^{k+1}\boldsymbol{\alpha}_j} = \arg\min_{\boldsymbol{\alpha}_j} \mathcal{L}({^k}\overline{\boldsymbol{\xi}}_j^*, {^k}\textbf{d}_j, \boldsymbol{\alpha}_j {^k}\boldsymbol{\lambda}_j, {^k}\textbf{s}_j ) \label{am_alpha}\\
        {^{k+1}\textbf{d}_j} = \arg\min_{\textbf{d}_j} \mathcal{L}({^k}\overline{\boldsymbol{\xi}}_j^*, \textbf{d}_j, {^{k+1}}\boldsymbol{\alpha}_j, {^k}\boldsymbol{\lambda}_j, {^k}\textbf{s}_j) \label{am_d} \\ 
        {^{k+1}}\mathbf{s}_j =\text{max}\left(0, -\mathbf{G} {^{k}}\overline{\boldsymbol{\xi}}_{j}^* - \textbf{y}_{lane}\right) \label{am_s} \\
        {^{k+1}}\boldsymbol{\lambda}_j = \overbrace{{^{k}}\boldsymbol{\lambda}_j+\rho\textbf{F}^T (\textbf{F}\hspace{0.05cm} {^k}\overline{\boldsymbol{\xi}}_j^*-{^{k}}\textbf{e}_j  )}^{\textbf{h}_1} \label{am_lambda} \\
        {^{k+1}}\textbf{e}_j = \overbrace{\begin{bmatrix}
        \widetilde{\textbf{e}} ({^{k+1}} \boldsymbol{\alpha}_j, {^{k+1}}\textbf{d}_j ) \label{am_e} \\
        \textbf{y}_{lane}-{^{k+1}}\textbf{s}_j
    \end{bmatrix}}^{\textbf{h}_2}\\
        {^{k+1}}\overline{\boldsymbol{\xi}}_j^* = \arg\min_{\overline{\boldsymbol{\xi}}_j^*}\mathcal{L}(\overline{\boldsymbol{\xi}}_j^*, {^{k+1}}\boldsymbol{\lambda}_j, {^{k+1}}\textbf{e}_j ) \label{am_xi}
    \end{align}
\end{subequations}
\normalsize

As can be seen, we optimize over only one group of variables at each AM step  while others are held fixed at values obtained at the previous updates. Steps \eqref{am_lambda}-\eqref{am_e} provides the function $\textbf{h}$ presented in \eqref{fixed_point_1}. That is, $\textbf{h} = (\textbf{h}_1, \textbf{h}_2)$. Step \eqref{am_xi} represents \eqref{fixed_point_2}. An important thing to note is that \eqref{am_alpha}, \eqref{am_d} have a closed-form solution in terms of ${^{k}}\overline{\boldsymbol{\xi}}^*_j$ and thus do not require any matrix factorization \cite{masnavi2022visibility}.

\vspace{-0.1cm}

\bibliography{ref_iros_2023}
\bibliographystyle{IEEEtran}

\end{document}